\documentclass[accepted]{uai2026} 
                        
\usepackage[american]{babel}

\usepackage{natbib} 
\usepackage{mathtools} 
\usepackage{booktabs} 
\usepackage{tikz} 
\usepackage{amsfonts, amssymb}
\usepackage{algpseudocode}
\usepackage{algorithm}
\usepackage{graphicx}
\usepackage{subcaption} 

\usepackage{ragged2e}
\usepackage{pgfplots}
\usepackage{pgfplotstable}
\usepgfplotslibrary{groupplots}
\pgfplotsset{compat=1.18}

\definecolor{ours}{RGB}{0,114,178}
\definecolor{ce}{RGB}{213,94,0}
\definecolor{sac}{RGB}{0,158,115}
\definecolor{naf}{RGB}{204,121,167}
\definecolor{mc}{RGB}{230,159,0}

\author[1,2]{\href{mailto:<shishir.sharma@mail.mcgill.ca>?Subject=Your UAI 2026 paper}{Shishir Sharma}{}}
\author[1,2]{Doina Precup}
\affil[1]{%
    McGill University, Montreal, Canada
}
\affil[2]{%
    Mila – Quebec Artificial Intelligence Institute, Montreal, Canada
}
\title{Analytic Planning under Uncertainty with Moment Closure}

\begin{document}
\maketitle

\begin{abstract}
Effective model-based reinforcement learning in stochastic environments requires planning that accounts for predictive uncertainty. Propagating full state distributions analytically offers a principled way to do this, but has traditionally required restrictive policy or reward structures to remain tractable. Consequently, modern deep reinforcement  learning
has largely retreated to either stochastic sampling, which introduces significant
target variance, or deterministic point estimates that ignore predictive
covariance entirely. We investigate whether distribution-aware planning is
possible without these constraints. Using a quadratic action-value 
parameterization, we first reduce the Bellman backup to an expectation over the
state-value function alone; the key idea is then a compatibility principle
between the predictive transition distribution and the value function class,
under which this expectation is analytic in the distribution's moments. We
instantiate this principle with a Gaussian transition model paired with a
radial-basis value function, yielding a closed-form backup that propagates both
predictive mean and covariance. Empirically, our approach reduces target variance
and yields well-calibrated predictive uncertainty under stochastic observations
in continuous control, providing a principled framework for planning with learned
distribution models.
\end{abstract}

\begin{figure*}[t]
    \centering
    \begin{subfigure}[t]{\linewidth}
        \centering
        \includegraphics[width=\linewidth]{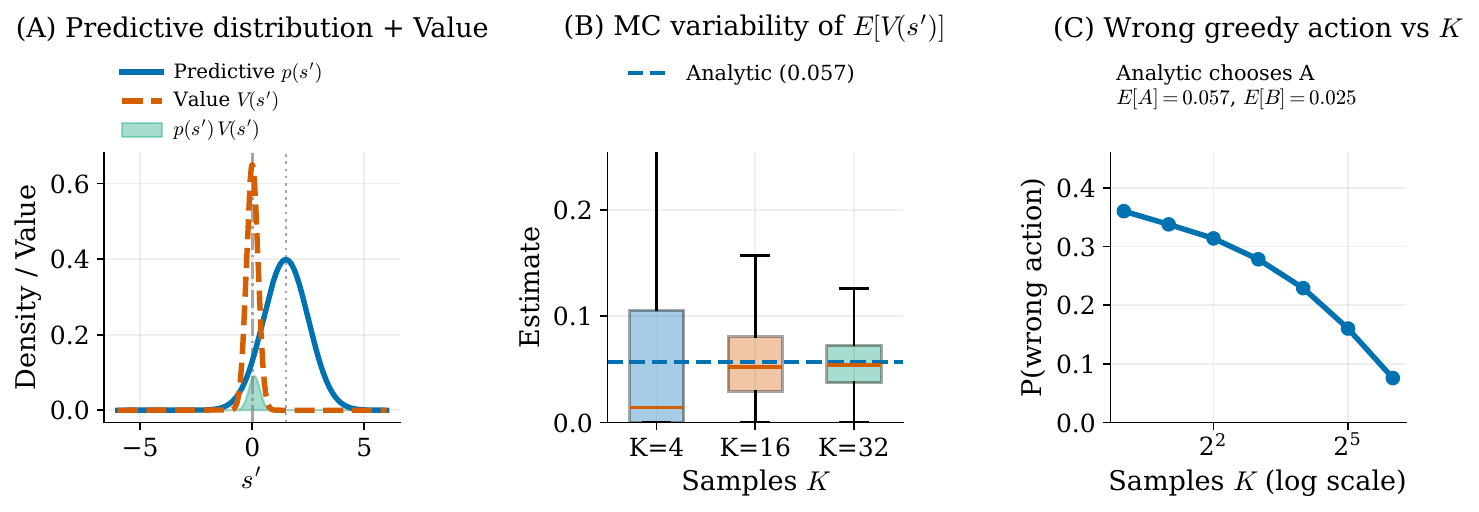}                
    \end{subfigure}
    \caption{%
        \textbf{Analytic versus Monte Carlo estimation of the Bellman
        expectation $\mathbb{E}[V(s')]$}, on a one-dimensional toy example with
        a Gaussian predictive distribution $p(s')$ and a radial-basis value function $V(s')$.
        \textbf{(A)}~The expectation is the integral of the density-weighted
        value $p(s')\,V(s')$ (shaded); Here equal to
        $\mathbb{E}[V(s')] = 0.057$.
        \textbf{(B)}~A Monte Carlo estimate from $K$ samples is unbiased but is high-variance, especially when $V$ is
        peaked. \textbf{(C)}~For a choice between two actions whose predictive means
        differ ($\mathbb{E}[A] = 0.057$, $\mathbb{E}[B] = 0.025$, so $A$ is
        better), sampling noise leads to the wrong action being chosen with
        higher probability at smaller $K$, whereas the analytic expectation
        always selects correctly.
    }
    \label{fig:toy}
\end{figure*}

\section{Introduction}
Recent advances in model-based reinforcement learning (MBRL) increasingly revolve around
\emph{world models} that enable imagination-based value learning and planning in either
state space or learned latent spaces \citep{gu2016continuous, chua2018deep,janner2019mbpo,hafner2019learning,Hafner2020Dream,hansen2022temporal,Hafner2025}.
A common pattern in these systems is that Bellman targets (or policy updates) are computed from
\emph{sampled} imagined rollouts or trajectory optimization under the learned dynamics, which
implicitly propagates predictive uncertainty through Monte Carlo variation.
While this approach scales to expressive models, it makes value estimation inherently stochastic and
often sensitive to sampling budgets and target variance.

Non-sampling approaches avoid this variance but have their own limitations.
Methods that collapse a stochastic model to its mean, or learn a deterministic
model directly, discard predictive covariance and can be systematically biased
when value functions are nonlinear. On the other hand, methods that propagate the
full predictive distribution analytically can often require strong structural 
constraints, for instance on the policy or reward, to remain tractable.

Our focus is distinct from both mean-substitution and rollout-style uncertainty propagation. 
Rather than propagating distributions over trajectories, we integrate stochastic one-step transitions directly inside the Bellman backup used for value-based control. The question we address is whether the Bellman expectation itself can be computed analytically under learned stochastic dynamics.

We show that, under a quadratic action-value parameterization, closed-form Bellman expectations are possible when the transition model and value representation are moment-compatible.  We instantiate this principle with a Gaussian dynamics model that predicts mean and covariance, paired with a value function whose expectation under a Gaussian distribution is analytic, resulting in a closed-form expression for the Bellman target. This removes Monte Carlo sampling from target construction while preserving propagation of predictive variance.

Concretely, we learn a heteroscedastic Gaussian transition model via maximum likelihood to capture aleatoric uncertainty and use an ensemble, combined via moment matching, to represent epistemic uncertainty. Greedy action selection is made tractable separately, via a quadratic action-value parameterization that removes the inner maximization from the Bellman backup; uncertainty then enters exclusively through the analytic Bellman expectation. The value function is parameterized as a mixture of Gaussian radial basis functions, whose expectation under the predictive Gaussian is available in closed form. Empirically, analytic greedy planning remains stable as observation noise increases, whereas sampling-based and mean-only baselines degrade. Ablations show that our moment-matched predictive distribution stably captures the combined uncertainty, and that this robustness stems from planning efficiently under that uncertainty rather than from increased Monte Carlo budgets.

\section{Related Work}
\label{sec:related}

MBRL under uncertainty has been approached from multiple directions, differing primarily in how stochastic dynamics influence value estimation and planning. Most contemporary deep MBRL systems estimate value targets by sampling imagined trajectories from learned dynamics models, using particle rollouts or model predictive control-style shooting methods \citep{chua2018deep,janner2019mbpo,hafner2019learning,Hafner2020Dream}. These approaches scale well and accommodate expressive neural models, but Monte Carlo sampling introduces variance directly into Bellman targets. Stability often depends on rollout depth, particle count, or model bias.

At the other end, expectation-style planning constructs targets through a deterministic dynamics representation — either by collapsing a stochastic model to its predicted mean, or by learning a deterministic model directly \citep{nagab2017neural,wan2019planning}. Such approaches simplify target computation and preserve determinism, but discard predictive covariance.

Probabilistic dynamics models and deep ensembles are widely used to represent aleatoric and epistemic uncertainty \citep{lakshminarayanan2017simple}. In online MBRL, ensembles mitigate model bias and improve exploration \citep{chua2018deep}. However, even when models are probabilistic, value targets are typically constructed either by sampling ensemble members or by collapsing ensemble predictions to point estimates.

Analytic uncertainty propagation offers a third perspective. In control and Gaussian process-based RL, moment matching has been used to propagate state distributions through nonlinear dynamics for policy search \citep{deisenroth2011pilco}. More recently, deterministic approximations have been proposed to reduce Monte Carlo noise in deep value updates by propagating Gaussian approximations through critics \citep{akgul2024deterministic}. These approaches typically approximate intermediate activations or rely on multi-step rollout evaluation.

Our work instead targets the Bellman backup itself in value-based deep MBRL, decoupling the tractability of greedy action selection from the tractability of the predictive expectation. Quadratic Q-function parameterizations enable closed-form greedy action selection in continuous control \citep{gu2016continuous}; we leverage this structure so that the maximization disappears from the backup entirely, leaving an expectation over the state-value function alone. This decoupling requires tractability only of the value representation, allowing analytic, uncertainty-aware planning without restricting the policy class or reward structure.

\section{Analytic Greedy Planning via Moment Closure}
\label{sec:method}

\subsection{Motivation}
\label{sec:motivation}

To illustrate the impact of sampling noise in Bellman backups, we construct a one-dimensional toy example.
Let the predictive distribution $p(s')$ be Gaussian and the value function
$V(s')$ be a single Gaussian radial basis function. The expectation is then
the integral of their pointwise product $p(s')\,V(s')$, shown shaded in
Figure~\ref{fig:toy}(A). When both $p$ and $V$ are Gaussian-shaped, this integral admits a
simple closed form, so the exact expectation is available without
any sampling.

A sampling-based planner instead draws $K$ next states
$s'_1, \dots, s'_K \sim p(s')$ and averages the value at those points,
$\tfrac{1}{K}\sum_{i} V(s'_i)$. This estimator is unbiased, but its variance
can be severe, especially when $V$ is sharply peaked relative to $p$, i.e. when only a narrow band of next states is
highly valued (imagine the edge of a cliff, where a small region of states is
safe and the rest are not). In this regime most sampled states fall where
$V \approx 0$ and contribute nothing, while the estimate is carried by the
rare samples that land near the peak. Figure~\ref{fig:toy}(B) shows the
consequence: at small $K$ the Monte Carlo estimate scatters widely around the
true expectation (dashed line), tightening only as $K$ grows.

This variance can easily corrupt control decisions. Greedy
action selection does not require the expected value itself, only the
\emph{ranking} of actions by expected value, and sampling noise can flip that
ranking. To show this, we consider two actions whose predictive distributions
differ only in their means, so that one action is genuinely better (higher
true expected value) than the other. A Monte Carlo planner selects whichever
action has the higher sampled average; Figure~\ref{fig:toy}(C) reports how
often this disagrees with the correct choice. At small $K$ the wrong action is
selected a substantial fraction of the time, and the error rate falls toward
zero only as $K$ increases. The analytic expectation, computed exactly, never
makes this error and removes this source of instability entirely.

\subsection{Preliminaries}
We consider the standard reinforcement learning setting formalized as a Markov decision process (MDP) \citep{sutton2018reinforcement}, defined by the tuple
\[
(\mathcal{S}, \mathcal{A}, p, r, \gamma),
\]
where $\mathcal{S} \subset \mathbb{R}^{d_s}$ is a continuous state space,
$\mathcal{A} \subset \mathbb{R}^{d_a}$ is a continuous action space,
$p(s'|s,a)$ is the transition function,
$r(s,a)$ is a bounded reward function,
and $\gamma \in [0,1)$ is a discount factor. For a policy $\pi$, the action-value function is
\[
Q^\pi(s,a)
=
\mathbb{E}_\pi
\left[
\sum_{k=0}^\infty \gamma^k r(S_{t+k},A_{t+k})
\;\middle|\;
S_t = s, A_t = a
\right]
\]
The optimal action-value function satisfies the Bellman optimality equation
\begin{equation}
Q^*(s,a)
=
\mathbb{E}_{s' \sim p(\cdot \mid s,a)}
\left[
r(s,a) + \gamma \max_{a' \in \mathcal{A}} Q^*(s',a')
\right]
\label{eq:bellman}
\end{equation}

\paragraph{Learned stochastic dynamics}
In MBRL, the true transition density $p(s'|s,a)$
is replaced by a learned model $\hat p_\theta(s'|s,a)$.
When $\hat p_\theta$ is stochastic, the Bellman target in~\eqref{eq:bellman}
requires computing an expectation over next states under the learned predictive distribution.

\subsection{Decoupling Greedy Action Selection from Expectation}
\paragraph{Representation challenge}
A sampling-free Bellman backup must compute
\[
\mathbb{E}_{s' \sim \hat p_\theta(\cdot \mid s,a)}
\left[
\max_{a' \in \mathcal{A}} Q(s',a')
\right]
\]
The difficulty arises because the Bellman expectation involves the greedy value
$\max_{a'} Q(s',a')$ evaluated at each possible next state $s'$.
Even if $Q(s',a')$ is smooth, the maximization over actions produces a highly nonlinear function of $s'$. As a result, even under Gaussian predictive dynamics, the Bellman expectation is generally intractable for expressive deep $Q$-functions.

One classical route to tractability is to impose strong structural assumptions. In linear--quadratic settings (LQR/LQG), the $Q$-function is quadratic jointly in the state and action, yielding closed-form greedy actions and analytic expectations under Gaussian disturbances \citep{anderson2007optimal}. Similarly, trajectory-optimization methods such as iterative LQG or differential dynamic programming maintain local quadratic approximations to the cost-to-go to obtain analytic feedback updates \citep{todorov2005generalized, tassa2012synthesis}. However, enforcing this joint quadratic structure in state and action directly on a deep Q-function would significantly restrict representational capacity.

\paragraph{Key observation: decoupling maximization from expectation}
The intractability of the Bellman expectation stems from integrating the greedy value
$\max_{a'} Q(s',a')$ under a stochastic transition.
If the maximization can be removed from inside the expectation, the backup reduces to evaluating
\[
\mathbb{E}_{s' \sim \hat p_\theta(\cdot \mid s,a)}[V(s')],
\]
which is substantially simpler.

We achieve this decoupling using a quadratic action-value parameterization (NAF-style) \citep{gu2016continuous}. Specifically, we decompose
\begin{equation}
Q(s,a) = V(s) - \frac{1}{2}(a-\tilde\mu(s))^\top P(s)(a-\tilde\mu(s)), \label{eq:naf_decomp}
\end{equation}
where $P(s)\succ 0$. Crucially, unlike the classical quadratic-in-state-and-action routes above, none of $V(s)$, $\tilde\mu(s)$, or $P(s)$ are restrained to be quadratic: only the action-slice at each fixed $s$ is quadratic, so $Q$ retains its full flexibility in how it varies with $s$. In the unconstrained case $\mathcal{A}=\mathbb{R}^{d_a}$, greedy action selection is available in closed form,

\[
\arg\max_a Q(s,a) = \tilde\mu(s), \qquad \max_a Q(s,a) = V(s).
\]

Substituting this structure into the Bellman equation removes the inner maximization entirely,
so that the backup value depends only on the state-value term:
\[
y(s,a)
=
r(s,a)
+
\gamma
\mathbb{E}_{s' \sim p(\cdot \mid s,a)}[V(s')]
\]

This decomposition has a consequence central to our approach. The value part $V(s)$ and the policy part $\tilde\mu(s)$ play separate roles: uncertainty enters
the backup only through the expectation of $V(s')$, so it is the value representation alone that must be made compatible with the predictive
distribution, while the greedy policy $\tilde\mu(s)$ carries no such compatibility requirement. In analytic planning with distribution models, the
two are typically tied together, with the policy class restricted so that the propagated state distribution stays tractable. Here the tractability requirement
falls entirely on the value representation, freeing the policy from any such constraint: we obtain analytic uncertainty propagation and an unrestricted,
network-parameterized policy at the same time.

\paragraph{Bounded actions}
In practical continuous-control settings, actions are typically bounded (e.g., $\mathcal{A}=[-1,1]^{d_a}$),
and the unconstrained maximizer $\tilde\mu(s)$ need not lie in $\mathcal{A}$.
Following the bounded NAF (BNAF) construction of \citet{plaksin2022continuous},
we define the greedy action as the projection
$\mu(s)=\Pi_{\mathcal{A}}(\tilde\mu(s))$
and shift the quadratic by a state-dependent constant,
\begin{align}
Q(s,a)
=
V(s)
&-\frac{1}{2}(a-\tilde\mu(s))^\top P(s)(a-\tilde\mu(s)) \nonumber \\
&+\frac{1}{2}(\mu(s)-\tilde\mu(s))^\top P(s)(\mu(s)-\tilde\mu(s))
\label{eq:bnaf_q}
\end{align}
This preserves quadratic structure while ensuring
\[
\arg\max_{a\in\mathcal{A}} Q(s,a) = \mu(s),
\qquad
\max_{a\in\mathcal{A}} Q(s,a) = V(s)
\]
Consequently, even under bounded actions, the Bellman backup reduces to computing $\mathbb{E}_{s' \sim p(\cdot \mid s,a)}[V(s')]$ such that uncertainty influences planning exclusively through the expectation of the state-value function.

For the class of RL problems arising from time-discretized
optimal control with control-affine dynamics and control-quadratic costs,
\citet{plaksin2022continuous} show that this bounded family is expressive
enough to solve the Bellman optimality equation to arbitrary accuracy as the
discretization is refined, and that any such approximate
solution yields a greedy policy with correspondingly bounded suboptimality. They further show that the original vertex-constrained NAF
family does not admit this guarantee, which motivates our use
of the bounded construction.

\subsection{Stochastic Dynamics Model}
\label{subsec:dynamics}

We learn a stochastic dynamics model as a conditional Gaussian,
\[
p_\theta(s' \mid s,a)
=
\mathcal{N}\!\big(s'; \mu_\theta(s,a), \Sigma_\theta(s,a)\big),
\]
where $\mu_\theta : \mathcal{S}\times\mathcal{A}\to\mathbb{R}^{d_s}$ predicts the conditional mean and
$\Sigma_\theta : \mathcal{S}\times\mathcal{A}\to\mathbb{R}^{d_s\times d_s}$ predicts the conditional covariance.
The covariance $\Sigma_\theta(s,a)$ captures \emph{aleatoric} uncertainty, corresponding to irreducible stochasticity in the environment.

\paragraph{Ensemble-based epistemic uncertainty}
To capture \emph{epistemic} uncertainty arising from limited data, instead of a single model, we train an ensemble of $K$ independently initialized copies
$\{(\mu_{\theta_k}, \Sigma_{\theta_k})\}_{k=1}^K$.
The ensemble induces a uniformly weighted mixture of Gaussian transition models,
\[
p(s' \mid s,a)
=
\frac{1}{K}
\sum_{k=1}^K
\mathcal{N}\!\big(s'; \mu_{\theta_k}(s,a), \Sigma_{\theta_k}(s,a)\big)
\]

Rather than sampling ensemble members during planning, we approximate this mixture by a single Gaussian whose mean and covariance match the first two moments of the mixture, following \citet{lakshminarayanan2017simple}.
Using the law of total expectation and total variance, the moment-matched mean is
\[
\bar\mu(s,a)
=
\frac{1}{K}
\sum_{k=1}^K
\mu_{\theta_k}(s,a),
\]
and the predictive covariance decomposes as
\begin{align}
\bar\Sigma(s,a)
={} & \frac{1}{K}\sum_{k=1}^K \Sigma_{\theta_k}(s,a)
\nonumber \\
{}+{} &  \frac{1}{K}\sum_{k=1}^K \Delta\mu_k(s,a)\,\Delta\mu_k(s,a)^\top,
\nonumber \\
\text{where }  \Delta\mu_k(s,a) ={} & \mu_{\theta_k}(s,a)-\bar\mu(s,a)
\label{eq:moment_match_cov}
\end{align}

The first term in $\bar\Sigma(s,a)$ is the ensemble-averaged aleatoric covariance from each member. The second term captures \emph{epistemic} uncertainty, reflecting divergence across ensemble members \citep{kendall2017uncertainties}. Because members are trained on independent bootstrap resamples, this divergence is a genuine disagreement signal, inflating the predictive covariance in regions where ensemble members disagree.

We therefore plan under the approximation
\[
p(s' \mid s,a)
\;\approx\;
\mathcal{N}\!\big(s'; \bar\mu(s,a), \bar\Sigma(s,a)\big),
\]
without sampling ensemble members.
Under this model, the Bellman target becomes
\begin{equation}
y(s,a)
=
r(s,a)
+
\gamma
\mathbb{E}_{s' \sim \mathcal{N}(\bar\mu(s,a),\bar\Sigma(s,a))}
\!\left[V(s')\right]
\label{eq:gaussian_backup}
\end{equation}
The remaining challenge is therefore to evaluate $\mathbb{E}[V(s')]$ in closed form.

\subsection{Moment-Compatible Value Representation}
\label{sec:mcvr}

The tractability of the analytic backup rests on a compatibility condition
between the predictive transition distribution and the value representation.
Let $p(s'| s,a)$ be the predictive distribution with parameters $\theta_p$.
We say the value class is \emph{moment-compatible} with $p$ if the expectation
$\mathbb{E}_{s' \sim p}[V(s')]$ admits a closed-form expression in $\theta_p$.
Whenever this holds, the state-value expectation in the Bellman backup can be
evaluated analytically, with no sampling.

Moment compatibility is a design principle rather than a prescription of any
single functional form. Several pairings of predictive 
distribution and value class could satisfy it, such as polynomial value functions with
elliptical distributions (e.g.\ Gaussian, Laplace, Student-$t$) or Random Fourier Features with a Gaussian
predictive distribution. For concreteness and simplicity, we parameterize the state-value function as a mixture of Gaussian radial basis functions,
\begin{equation}
V(s)
=
\sum_{i=1}^M
w_i
\exp\!\left(
-\frac{1}{2}
(s - c_i)^\top
\Lambda_i
(s - c_i)
\right),
\label{eq:rbf_value}
\end{equation}
where $\Lambda_i \succ 0$. The centers $c_i$ and precision matrices $\Lambda_i$ are fixed throughout training (see Appendix~\ref{app:init}-\ref{app:init2} for initialization details); only the weights $\{w_i\}$ are learned, updated via temporal-difference learning as part of the critic optimization (Algorithm \ref{alg:mcmb}, line~12).

Under a Gaussian next-state distribution $s' \sim \mathcal{N}(\mu,\Sigma)$, the
expectation of each basis function can be computed exactly. Writing
$\|x-a\|_B^2 := (x-a)^\top B\,(x-a)$ and $A_i := \Lambda_i(I + \Sigma\Lambda_i)^{-1}$,
\begin{equation}
\mathbb{E}[V(s')]
= \sum_{i=1}^{M} w_i\, |I + \Sigma\Lambda_i|^{-1/2}
  \exp\!\big(-\tfrac12 \|\mu - c_i\|_{A_i}^2\big)
\label{eq:analytic_expectation}
\end{equation}
The full derivation is given in Appendix~\ref{app:gaussian_expectation}.

\begin{algorithm}[ht]
\caption{\textbf{Mo}ment-\textbf{C}ompatible \nobreak \textbf{A}nalytic Planning }
\label{alg:mcmb}
\begin{algorithmic}[1]
\raggedright
\State Initialize twin critics $\phi_1,\phi_2$ (each with $V_\phi,\mu_\phi$) and target parameters $\phi_1^-,\phi_2^-$, ensemble dynamics models $\{p_{\theta_k}\}_{k=1}^K$, replay buffer $\mathcal{D}$, update counter $n\gets 0$
\For{each environment step}
    \State With probability $p$, $a\sim\mathrm{Uniform}(\mathcal{A})$; otherwise $a = \Pi_{\mathcal{A}}(\mu_{\phi_1}(s) + \epsilon)$, $\epsilon\sim\mathcal{N}(0,\sigma^2 I)$ \Comment{exploration}
    \State Execute $a$, observe $(r,s',d)$, and store $(s,a,r,s',d)$ in $\mathcal{D}$
    \For{each gradient step}
        \State Sample minibatch $\mathcal{B}\subset\mathcal{D}$
        \For{$k=1$ to $K$}
            \State Update dynamics parameters $\theta_k$ via NLL on bootstrap resample of $\mathcal{B}$
        \EndFor
        \State Compute moment-matched predictive Gaussian $(\bar\mu,\bar\Sigma)$ from $\{(\mu_{\theta_k},\Sigma_{\theta_k})\}_{k=1}^K$ using Eq.~\eqref{eq:moment_match_cov}
        \State Compute target $y \gets r + \gamma(1-d)\min_{i=1,2}\mathbb{E}_{s'\sim \mathcal{N}(\bar\mu,\bar\Sigma)}[V_{\phi_i^-}(s')]$ 
        \State Update critics $\phi_1,\phi_2$ by minimizing $(Q_{\phi_i}(s,a)-y)^2$ on $\mathcal{B}$
        \State $n\gets n+1$; every $N_\text{target}$ updates, $\phi_i^-\gets\phi_i$
    \EndFor
\EndFor
\end{algorithmic}
\end{algorithm}

\section{Learning Procedure}
\label{sec:learning}

We introduce \textbf{Mo}ment-\textbf{C}ompatible \nobreak \textbf{A}nalytic Planning (MoCA) and detail the overall training procedure (Algorithm \ref{alg:mcmb}). We integrate the analytic Bellman backup into a standard off-policy learning loop with a learned stochastic dynamics ensemble. The agent alternates between collecting transitions, updating the dynamics model, and updating value critics using analytic targets.

\paragraph{Exploration}
Actions are selected from the greedy policy implied by the quadratic critic, with injected exploration noise.
Concretely, we perturb the greedy action with additive Gaussian noise and clip to the feasible action set.
In addition, with a fixed probability we instead take a uniformly random action (also clipped).
This mirrors the exploration mechanism used in Spinning Up’s TD3 implementation \citep{SpinningUp2018}, combining local exploration around the greedy action with occasional global exploration.

\paragraph{Dynamics model}
Given a replay buffer $\mathcal{D} = \{(s_t,a_t,s_{t+1},r_t,d_t)\}$, 
we train each member of a Gaussian dynamics ensemble
\[
s' \sim \mathcal{N}(\mu_{\theta_k}(s,a), \Sigma_{\theta_k}(s,a))
\]
by minimizing the negative log-likelihood.
Models are trained independently on bootstrap resamples of the sampled minibatch $\mathcal{B}$.
At target-construction time, ensemble predictions are aggregated via moment matching according to~\eqref{eq:moment_match_cov}, yielding a single predictive Gaussian
$\mathcal{N}(\bar\mu(s,a),\bar\Sigma(s,a))$.

\paragraph{Twin value critics and analytic targets}
To mitigate overestimation bias, we adopt clipped double estimation (as in TD3), adapted to our setting where greedy control reduces the backup to a state-value term.
We maintain two critics, each inducing a state-value function $V_{\phi_1}$ and $V_{\phi_2}$ (with corresponding target parameters $\phi_1^-,\phi_2^-$).
For a transition $(s_t,a_t,r_t,d_t)$, we compute two analytic expected next-values under the moment-matched predictive Gaussian:
\[
v_i
=
\mathbb{E}_{s' \sim \mathcal{N}(\bar\mu(s_t,a_t), \bar\Sigma(s_t,a_t))}
\!\left[ V_{\phi_i^-}(s') \right],
\quad i \in \{1,2\},
\]
where the expectation is evaluated in closed form via $V$'s moment-compatible representation (Section~\ref{sec:mcvr}).
The Bellman target then uses clipped double estimation:
\[
y_t
=
r_t
+
\gamma(1-d_t)\,\min(v_1,v_2)
\]
Each critic is updated by minimizing a squared-error regression to this shared target.
Target parameters are refreshed periodically by copying the online critic parameters.

\paragraph{Action bounds and model evaluation}
A related and well-known failure mode in model-based RL is \emph{model exploitation} (or \emph{planner overfitting}), where planning discovers actions that exploit inaccuracies of the learned dynamics in poorly supported regions of the state--action space, leading to overly optimistic value estimates.
In our setting, this risk is particularly transparent because analytic expectation removes Monte Carlo variability from target construction: any optimism induced by off-support model queries is no longer masked by sampling noise and can appear as systematic bias.
For this reason, we restrict model evaluation to observed $(s,a)$ pairs when constructing Bellman targets and avoid Dyna-style synthetic rollouts; greedy actions are projected onto the feasible action set via the BNAF construction, and uncertainty enters only through the closed-form expectation of $V(s')$.

If one were to extend the method to Dyna-style synthetic experience generation, the BNAF correction term in~\eqref{eq:bnaf_q} would need to be incorporated inside the expectation.
Analytic treatment under bounded actions would likely require truncated Gaussian moments, which we leave for future work.

\begin{figure*}[t]
\centering
\includegraphics[width=\linewidth]{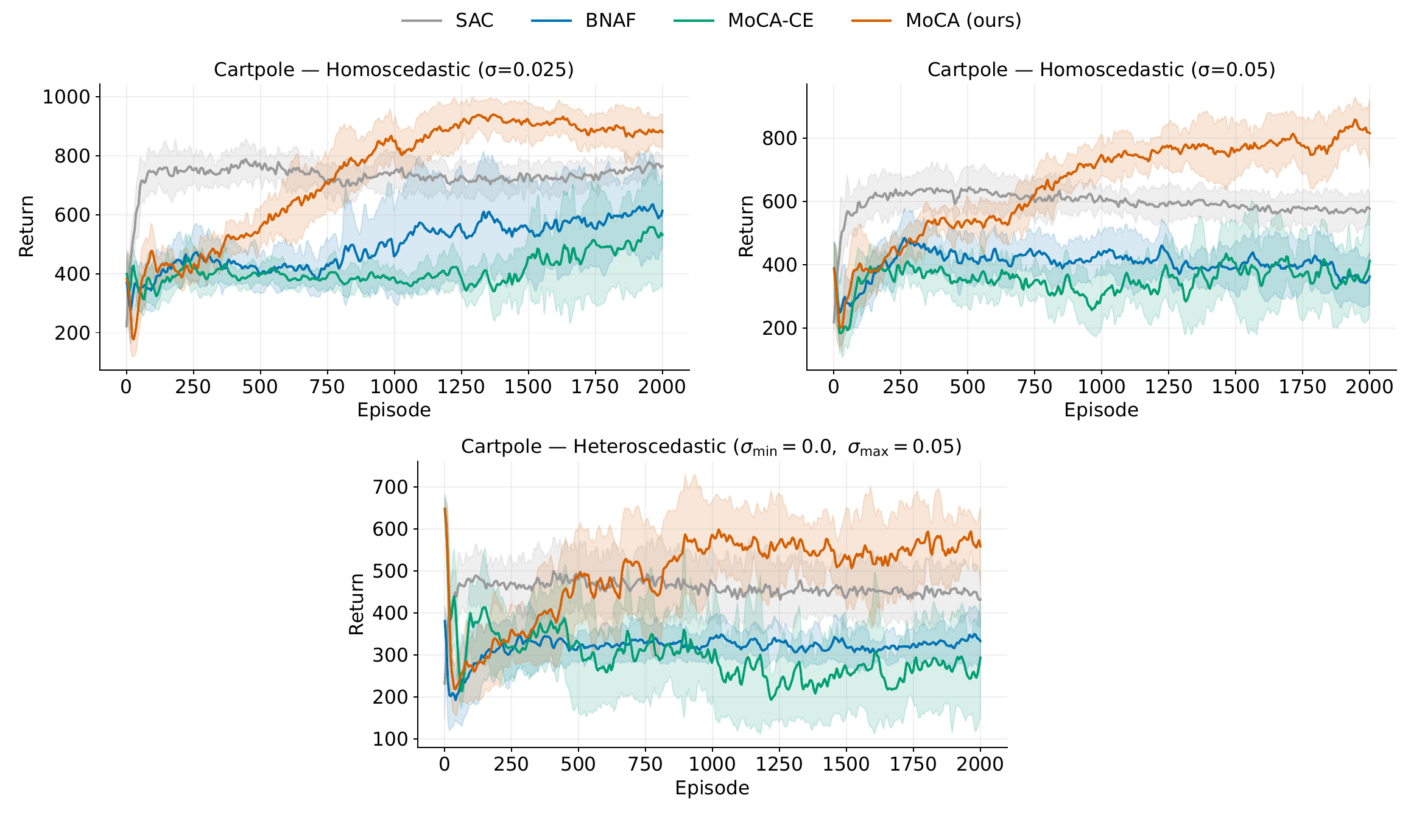}
\caption{Learning curves on Cartpole under Gaussian observation noise. Top row: homoscedastic noise with standard deviation $\sigma \in {0.025, 0.05}$. Bottom: heteroscedastic noise with $\sigma_{\min}=0.0$, $\sigma_{\max}=0.05$. Curves show mean return over 10 seeds; shaded regions denote $\pm$1 standard deviation.}
\label{fig:Cartpole}
\end{figure*}

\begin{figure*}[t]
\centering
\includegraphics[width=\linewidth]{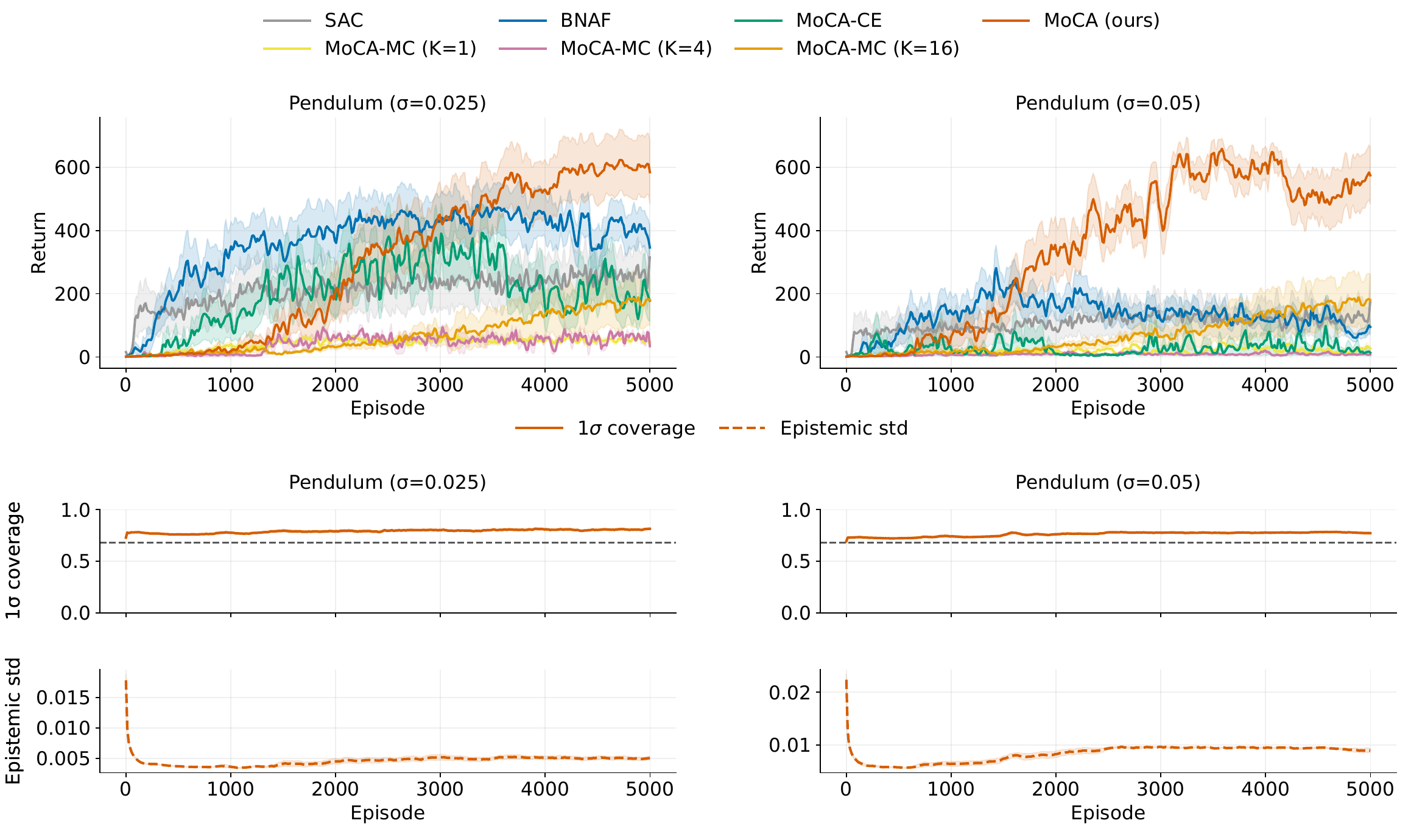}
\caption{Return and uncertainty diagnostics on Pendulum under Gaussian observation noise ($\sigma \in \{0.025, 0.05\}$, left and right columns). Top: evaluation return during training. Middle: for MoCA, empirical $1\sigma$ coverage remains near the nominal 68\% level throughout training. Bottom: epistemic standard deviation for MoCA, which stabilizes as the ensemble converges.}
\label{fig:Pedulum_homo}
\end{figure*}

\section{Empirical Study}
\label{sec:experiments}

\subsection{Implementation Details}

Our model-free baseline follows the continuous Normalized Advantage Function (NAF) formulation of \citet{gu2016continuous} and its bounded-action analysis in \citet{plaksin2022continuous}. We adopt the faithful bounded variant (BNAF), in which the quadratic vertex $\tilde{\mu}(s) \in \mathbb{R}^A$ is learned without squashing and only clipped to the environment action limits at evaluation time. The $Q$-function is parameterized so that $Q(s,\pi(s)) = V(s)$ for the clipped greedy action $\pi(s)=\mathrm{clip}(\tilde{\mu}(s))$, eliminating the saturation pathologies that arise from naïve tanh-constrained quadratic parameterizations.

The critic uses a shared MLP trunk with two heads: (i) an unconstrained vertex head $\tilde{\mu}(s)$ and (ii) a Cholesky head $L(s)$ defining a positive-definite curvature matrix $P(s)=L(s)L(s)^\top$. The value function $V(s)$ is parameterized linearly in fixed radial basis features. To ensure strict positive definiteness and prevent curvature collapse during training, we enforce a minimum-curvature constraint on $P(s)$ via the Cholesky parameterization.\footnote{We also apply a curvature keep-alive penalty that discourages near-flat quadratic advantages by adding $\lambda\,\mathbb{E}[\mathrm{ReLU}(p_{\min}-\|L(s)\|_F^2)^2]$ to the TD loss, where $\|L\|_F^2=\mathrm{tr}(P)$. See Appendix~\ref{app:curvature_reg}.}

Training uses twin critics with clipped double targets, i.e., Bellman targets are constructed using $\min\{V_1^{-}(s'), V_2^{-}(s')\}$ from periodically updated target networks. Optimization is performed with Adam on minibatches sampled from a replay buffer.

\subsection{Noise Models}
\label{sec:noise_models}
Our objective is to evaluate whether explicitly modeling predictive uncertainty improves planning performance under stochastic observations. To isolate the effect of uncertainty propagation in value estimation, we introduce controlled observation noise processes while keeping the underlying system dynamics unchanged.

A key consideration is avoiding transition-model misspecification. Since our learned dynamics model assumes Gaussian structure, we restrict injected stochasticity to Gaussian observation noise. This ensures that performance differences reflect how uncertainty is represented and propagated in planning, rather than mismatches in the transition distribution.

We consider two observation-noise regimes.

\paragraph{Homoscedastic Gaussian Observation Noise}

We first inject independent Gaussian noise directly into the observations:
\begin{equation}
\tilde{s}_t = s_t + \epsilon_t, 
\qquad 
\epsilon_t \sim \mathcal{N}(0, \sigma^2 I)
\end{equation}

The underlying dynamics remain deterministic; only the agent’s perception is corrupted. The variance is constant across the state space, providing a controlled and analytically simple stochastic regime. By varying $\sigma$, we obtain degradation curves that quantify robustness as observation uncertainty increases.

This homoscedastic setting serves as our primary evaluation regime, as it cleanly isolates the effect of uncertainty-aware planning under well-specified Gaussian noise.

\paragraph{State-Norm–Scaled (Heteroscedastic) Observation Noise}

Many practical systems exhibit observation uncertainty that increases with state magnitude. To reflect this structure, we consider state-dependent Gaussian observation noise of the form
\begin{equation}
\tilde{s}_t = s_t + \epsilon_t, 
\qquad 
\epsilon_t \sim \mathcal{N}\!\left(0, \sigma(s_t)^2 I\right),
\end{equation}
where the standard deviation scales with a normalized state norm:
\begin{equation}
\sigma(s_t) 
= \sigma_{\min} 
+ (\sigma_{\max} - \sigma_{\min})
\, \mathrm{clip}\!\left(
\frac{
\left\| \frac{s_t}{s_{\max}} \right\|_2
}{
m_{\max}
},
0, 1
\right)
\end{equation}

Here $s_{\max} \in \mathbb{R}^d$ is a fixed per-dimension scaling vector, and the division is understood elementwise. The quantity $m_{\max}$ bounds the normalized norm, ensuring that $\sigma(s_t) \in [\sigma_{\min}, \sigma_{\max}]$. 

Under this model, observation variance increases smoothly as the state moves away from nominal regions, capturing a common phenomenon in control systems where sensor uncertainty grows in high-magnitude or dynamically extreme regimes.

To maintain experimental clarity and controlled comparisons, our primary ablations focus on the homoscedastic regime, while the state-norm noise serves as a practically motivated extension that stresses the planner under structured, state-dependent uncertainty.

\subsection{Evaluation Protocol}

We evaluate all methods on continuous-control tasks from the MuJoCo Playground \cite{zakka2025mujocoplayground}, which provides JAX-based implementations of DeepMind Control Suite environments \cite{tassa2018deepmindcontrolsuite}. We report results on \emph{Cartpole Balance} and \emph{Pendulum Swingup}.

These environments do not contain terminal states; we therefore use fixed-length episodes of 1000 timesteps for both training and evaluation. After each training episode, the current policy is evaluated over 10 episodes without exploration noise, and the average return is recorded. All methods use identical interaction budgets, episode lengths, and update-to-data ratios. Results are averaged over ten random seeds.

On \emph{Cartpole Balance}, we study robustness to observation noise. We consider homoscedastic Gaussian noise with $\sigma \in \{0.025, 0.05\}$ and state-norm–scaled heteroscedastic noise with $\sigma_{\min}=0.0$ and $\sigma_{\max}=0.05$ (Section~\ref{sec:noise_models}). We compare Soft-Actor Critic (SAC) \citep{haarnoja2018soft} and BNAF against our analytic uncertainty-propagation method (MoCA) as well as a certainty-equivalent version (MoCA-CE), which computes the Bellman target using only the predictive mean $\bar\mu$ (i.e., $\bar\Sigma \to 0$ in Eq.~\eqref{eq:analytic_expectation}).

On \emph{Pendulum Swingup}, we isolate analytic versus Monte Carlo uncertainty propagation under homoscedastic Gaussian observation noise ($\sigma \in \{0.025, 0.05\}$). In addition to SAC, BNAF, and MoCA-CE, we include Monte Carlo variants MoCA-MC with $K=4$ and $K=16$ samples, and compare them against MoCA, which computes the same expectation analytically. We also report empirical $1\sigma$ coverage and epistemic standard deviation for MoCA to assess calibration during training.

\section{Results}

We conduct two complementary experiments.
First, we evaluate robustness to stochastic observations under controlled Gaussian noise regimes.
Second, we isolate the effect of analytic versus Monte Carlo uncertainty propagation in the Bellman expectation through a targeted ablation study.

\subsection{Robustness to Stochastic Observations}

Figure \ref{fig:Cartpole} reports learning curves on Cartpole under increasing observation noise.

\paragraph{Homoscedastic noise}
Under moderate noise ($\sigma=0.025$), all methods learn, but clear performance differences emerge. SAC and BNAF achieve reduced asymptotic return relative to low-noise settings, and MoCA-CE performs similarly to BNAF. In contrast, MoCA consistently attains higher final return and exhibits more stable learning dynamics.

As noise increases ($\sigma=0.05$), the gap widens substantially. SAC, BNAF, and MoCA-CE degrade further in both convergence speed and asymptotic performance. MoCA remains comparatively stable and maintains a clear performance advantage.

\paragraph{State-dependent noise}
Under state-norm–scaled observation noise, the same qualitative ordering persists. All baselines deteriorate noticeably, while MoCA retains superior stability and final return. Overall, the performance gains are attributable to analytic propagation of predictive uncertainty, rather than to model-based learning alone.

\subsection{Analytic vs.\ Monte Carlo Uncertainty Propagation}

Figure \ref{fig:Pedulum_homo} compares analytic and sampling-based uncertainty propagation on \emph{Pendulum Swingup}.

Under moderate noise, MoCA consistently achieves the highest return and fastest convergence. 
MoCA-MC improves as the number of samples increases from $K=4$ to $K=16$, but remains significantly below the analytic method in both learning speed and asymptotic performance. 
MoCA-CE and model-free baselines perform substantially worse, indicating that ignoring predictive variance is detrimental under stochastic observations.

As noise increases, these differences become more pronounced. 
Monte Carlo variants exhibit higher variance and slower improvement, even with $K=16$ samples. 
In contrast, MoCA maintains stable learning dynamics and a clear performance advantage.

The lower panels report uncertainty diagnostics for MoCA. 
Empirical $1\sigma$ coverage remains close to the nominal 68\% level throughout training, while the epistemic standard deviation stabilizes as learning progresses. 
This indicates that the analytic moment propagation remains well-calibrated while enabling superior control performance.

Overall, analytic uncertainty propagation provides both improved sample efficiency and greater robustness compared to stochastic Monte Carlo estimation under identical model assumptions.

\section{Discussion}

\paragraph{The role of each structural assumption}
Our method combines three structural choices that differ in both origin and function. Two are inherited from prior work: modeling the transition kernel as a Gaussian (mixture) is standard in MBRL \citep{chua2018deep, janner2019mbpo}, and condensing the ensemble mixture to a single moment-matched Gaussian follows \citet{lakshminarayanan2017simple}. The third, the radial-basis value class, is the element we introduce. It is worth separating what each assumption is for. The quadratic action-value (BNAF) parameterization is a structural necessity of the
framework: without a closed-form $\max_a Q$, the Bellman backup cannot be posed as
a state-value expectation in the first place. The dynamics-value pairing plays a different role. It is what
makes the resulting expectation $\mathbb{E}[V(s')]$ tractable, and the specific
Gaussian--RBF pair is one of several pairings that could satisfy this requirement.

\paragraph{Limitations of the value representation}
The radial-basis instantiation carries two distinct limitations in higher
dimensions. The first is representational: RBF mixtures are universal
approximators on compact state domains, but the number of bases $M$ needed to
maintain approximation quality grows poorly with state dimension, a manifestation
of the curse of dimensionality. The second is computational: for
each basis, the closed-form expectation requires the determinant
$|I + \Sigma\Lambda_i|^{-1/2}$ and the inverse $(I + \Sigma\Lambda_i)^{-1}$ of a
$d \times d$ matrix, each $O(d^3)$ in general, giving a per-backup cost of
$O(M d^3)$. When $\Lambda_i$ and the predictive covariance are diagonal, as in our
experiments, $I + \Sigma\Lambda_i$ is diagonal and both reduce to elementwise
operations, lowering the cost to $O(M d)$. One way to scale to higher dimensions could be to apply the analytic backup in a learned
lower-dimensional latent space, where a compact value representation suffices. A full
exploration of this direction is left for future work.
\section{Conclusion}
\label{sec:conclusion}

We introduced a moment-closure formulation for model-based reinforcement learning
that enables analytic Bellman expectations under learned stochastic dynamics. Two
ingredients combine to make this possible. First, a quadratic action-value
parameterization removes the inner maximization from the Bellman backup, reducing
it to an expectation of the state-value function alone. Second, a
moment-compatibility principle: whenever the predictive transition distribution
and the value representation are chosen so that this remaining state-value
expectation is closed-form in the distribution's moments, the backup can be
evaluated analytically, without sampling. This principle is not tied to any single
functional form. We realize it concretely with a Gaussian transition model paired
with a radial-basis value function, for which the expectation has a closed form
that propagates both predictive mean and covariance while remaining deterministic
and computationally efficient.

Empirically, analytic uncertainty propagation improves robustness to stochastic observations and consistently outperforms both mean-only (certainty-equivalent) and Monte Carlo baselines. In particular, it yields higher final performance, greater stability under increasing noise, and well-calibrated predictive uncertainty. These results demonstrate that analytic moment propagation can serve as a practical and effective alternative to sampling-based planning in continuous-control settings.

\bibliography{uai2026-template}
\clearpage

\appendix

\section{Observation Noise Construction}
\label{app:noise_details}

This appendix provides full implementation details of the observation noise processes used in Section \ref{sec:noise_models}.

\subsection{Homoscedastic Gaussian Noise}

In the homoscedastic setting, independent Gaussian noise is added directly to the observed state:
\begin{equation}
\tilde{s}_t = s_t + \epsilon_t, 
\qquad 
\epsilon_t \sim \mathcal{N}(0, \sigma^2 I). \nonumber
\end{equation}

The underlying dynamics remain deterministic; stochasticity enters only through the agent’s observations.
The variance $\sigma^2$ is fixed globally and identical across all state dimensions.
We evaluate $\sigma \in \{0.025, 0.05\}$.

\subsection{State-Norm–Scaled (Heteroscedastic) Noise}

To model state-dependent sensing uncertainty, we inject Gaussian noise with variance scaled by a normalized state magnitude.

Noise is defined as:
\begin{equation}
\tilde{s}_t = s_t + \epsilon_t,
\qquad
\epsilon_t \sim \mathcal{N}\!\left(0, \sigma(s_t)^2 I\right), \nonumber
\end{equation}
where the standard deviation is
\begin{equation}
\sigma(s_t)
=
\sigma_{\min}
+
(\sigma_{\max} - \sigma_{\min})
\cdot
\mathrm{clip}
\left(
\frac{\left\| s_t \oslash s_{\max} \right\|_2}{m_{\max}},
\, 0, 1
\right)
\label{eq:state_norm_noise}
\end{equation}

Here:

\begin{itemize}
\item $\oslash$ denotes elementwise division.
\item $s_{\max} \in \mathbb{R}^k$ is a fixed per-dimension scaling vector over the $k$ normalized state dimensions (defined below).
\item $m_{\max}$ bounds the normalized norm.
\item $\mathrm{clip}(\cdot,0,1)$ truncates the scaling factor to the interval $[0,1]$.
\end{itemize}

\paragraph{Per-Dimension Scaling}

We select $k=3$ state dimensions to include in the norm --- cart position, cart velocity, and pole angular velocity --- and exclude $\cos\theta,\sin\theta$, since these are already bounded by construction. Each selected dimension is first normalized by a fixed constant reflecting its nominal magnitude:
\begin{equation}
\hat{s}_t = s_t \oslash s_{\max}.
\end{equation}

For Cartpole, we use:
\begin{itemize}
\item cart position scale: $2.4$
\item cart velocity scale: $2.0$
\item pole angular velocity scale: $3.0$
\end{itemize}

These constants ensure all normalized state components are approximately $\mathcal{O}(1)$ within the typical operating region.

\paragraph{Normalized Magnitude}

The scaling term uses the (unsquared) $\ell_2$ norm:
\begin{equation}
m(s_t)
=
\frac{\|\hat{s}_t\|_2}{m_{\max}}.
\end{equation}

We set $m_{\max} = \sqrt{k}$, where $k$ is the number of normalized dimensions, such that $m(s_t) \approx 1$ when each normalized component has magnitude 1. For Cartpole, $k=3$, so $m_{\max} = \sqrt{3} \approx 1.732$. This produces smooth growth of observation variance as the system moves toward dynamically extreme regimes.

\paragraph{Hyperparameters}

For the heteroscedastic regime we use $\sigma_{\min} = 0.0, \sigma_{\max} = 0.05$.

\section{Moment-Compatible RBF Value Representation}
\label{app:rbf_details}

To enable analytic Bellman expectations under Gaussian predictive dynamics, the state-value function must admit a closed-form expectation under a Gaussian distribution. We therefore parameterize the value function as a finite mixture of Gaussian radial basis functions (RBFs).

\subsection{Functional Form}

The state-value function is defined as:
\begin{equation}
V(s)
=
\sum_{i=1}^{M}
w_i
\exp
\left(
-\frac{1}{2}
(s - c_i)^\top
\Lambda_i
(s - c_i)
\right),
\end{equation}
where:
\begin{itemize}
    \item $M$ is the number of RBF components,
    \item $c_i \in \mathbb{R}^d$ are fixed centers,
    \item $\Lambda_i \succ 0$ are positive-definite precision matrices,
    \item $w_i \in \mathbb{R}$ are learned weights.
\end{itemize}

The weights $\{w_i\}$ are optimized via temporal-difference learning. The centers and precision matrices are fixed throughout training.

\subsection{Gaussian Expectation}
\label{app:gaussian_expectation}
If the predictive next-state distribution is Gaussian, $s' \sim \mathcal{N}(\mu,\Sigma)$,
the expectation of each basis function admits a closed-form expression. For
brevity, write the quadratic form
\begin{equation}
\|x - a\|_B^2 := (x-a)^\top B\, (x-a) \nonumber
\end{equation}
Consider a single component with center $c_i$ and precision $\Lambda_i$:
\begin{equation}
\mathbb{E}[V_i(s')]
= w_i \int \exp\!\big(-\tfrac12 \|s'-c_i\|_{\Lambda_i}^2\big)\,
  \mathcal{N}(s';\mu,\Sigma)\, ds' \nonumber
\end{equation}
Writing the Gaussian density explicitly, the integrand is the exponential of a
quadratic form in $s'$,
\begin{equation}
\begin{split}
& Z_i \int \exp\!\big(-\tfrac12 \big[\, \|s'-c_i\|_{\Lambda_i}^2
  + \|s'-\mu\|_{\Sigma^{-1}}^2 \,\big]\big)\, ds', \nonumber \\
& \text{where } Z_i := w_i (2\pi)^{-d/2}|\Sigma|^{-1/2}
\end{split}
\end{equation}
Completing the square in $s'$ combines the two quadratic forms into a single
Gaussian with precision $\Lambda_i + \Sigma^{-1}$, leaving a constant factor that
depends on $\mu$ and $c_i$. Carrying out the Gaussian integral and simplifying
with the identity
$\Sigma^{-1}(\Lambda_i + \Sigma^{-1})^{-1}\Lambda_i = \Lambda_i(I + \Sigma\Lambda_i)^{-1}$
yields
\begin{align}
\mathbb{E}[V_i(s')]
=& \; w_i\, |I + \Sigma\Lambda_i|^{-1/2}
  \exp\!\big(-\tfrac12 \|\mu - c_i\|_{A_i}^2\big), \nonumber \\
\quad A_i :=& \; \Lambda_i(I + \Sigma\Lambda_i)^{-1} \nonumber
\end{align}
Summing over the $M$ components gives the full expectation,
\begin{equation}
\mathbb{E}[V(s')]
= \sum_{i=1}^{M} w_i\, |I + \Sigma\Lambda_i|^{-1/2}
  \exp\!\big(-\tfrac12 \|\mu - c_i\|_{A_i}^2\big) \nonumber
\end{equation}
which is Eq.~(7). This closed form enables deterministic Bellman targets without
Monte Carlo sampling.

\subsection{Number of Basis Functions}

We set the number of RBFs based on the effective complexity of the value landscape in each environment:
\begin{itemize}
    \item \textbf{Cartpole:} $M=256$ bases. Cartpole uses a 5D observation with two unbounded-velocity dimensions (practically bounded for initialization), producing a broader range of relevant regimes; we therefore allocate more bases to avoid underfitting. 
    \item \textbf{Pendulum:} $M=128$ bases. Pendulum observations are 3D with $(\cos\theta,\sin\theta)$ bounded by construction, making the value function smoother over a lower-dimensional bounded domain, requiring fewer bases.
\end{itemize}
These choices follow the principle that $M$ should scale with (i) state dimension and (ii) the extent of the region in which accurate value curvature is needed for stable greedy control.

\subsection{Center Initialization and Support Bounds}
\label{app:init}
RBF centers $\{c_i\}_{i=1}^M$ are sampled uniformly over a fixed hyper-rectangle defined by environment-specific practical bounds:
\[
c_i \sim \mathcal{U}(\ell, h),
\]
where $(\ell,h)$ are per-dimension lower/upper bounds.

\paragraph{Cartpole bounds}
We use bounds
\begin{align}
&\ell = [-2.5,\,-1.0,\,-1.0,\,-10.0,\,-15.0],\quad \nonumber \\
&h = [ 2.5,\, 1.0,\, 1.0,\, 10.0,\, 15.0],
\end{align}
corresponding to practical operating ranges for stable initialization, while $\cos\theta$ and $\sin\theta$ remain bounded by construction.

\paragraph{Pendulum bounds}
We use bounds
\[
\ell = [-1.0,\,-1.0,\,-8.0],\quad
h = [ 1.0,\, 1.0,\, 8.0].
\]

\subsection{Anisotropic Lengthscales}
\label{app:init2}
Each basis uses a diagonal precision (anisotropic bandwidth) to reflect heterogeneous units and curvature across state dimensions:
\[
\Lambda = \mathrm{diag}\left(\lambda_1^{-2},\dots,\lambda_d^{-2}\right).
\]

\paragraph{Cartpole lengthscales}
We set
\[
\lambda = [0.6,\,0.5,\,0.5,\,3.0,\,4.0],
\]
using tighter scales for $(\cos\theta,\sin\theta)$ and looser scales for velocity-related dimensions.

\paragraph{Pendulum lengthscales}
We set
\[
\lambda = [0.5,\,0.5,\,2.0],
\]
again using tighter scales on bounded trigonometric coordinates and a looser scale for angular velocity.

\section{Curvature Regularization (Keep-Alive)}
\label{app:curvature_reg}

For NAF-based critics, the advantage term uses a positive-definite curvature matrix $P(s)=L(s)L(s)^\top$, where $L(s)$ is produced by a Cholesky head. In addition to the Cholesky parameterization, we optionally apply a \emph{curvature keep-alive} penalty that prevents the quadratic form from becoming degenerate (near-flat) early in training.

\paragraph{Keep-alive penalty}
We add the following term to the critic TD loss on each replay minibatch:
\begin{equation}
\mathcal{L}_{\text{keepalive}}
=
\lambda \;
\mathbb{E}_{s \sim \mathcal{B}}
\left[
\mathrm{ReLU}\!\left(p_{\min} - \|L(s)\|_F^2\right)^2
\right],
\label{eq:keepalive}
\end{equation}
where $\mathcal{B}$ denotes the replay batch and $\|L(s)\|_F^2 = \sum_{i,j} L_{ij}(s)^2$.

\paragraph{Interpretation}
Since $\mathrm{tr}(P(s)) = \|L(s)\|_F^2$, Eq.~\eqref{eq:keepalive} enforces a lower bound on the \emph{overall curvature magnitude} (trace of $P$), rather than directly constraining individual eigenvalues. This reduces training pathologies where the learned quadratic advantage becomes nearly flat, which can destabilize greedy control and TD learning.

\paragraph{Scope}
This keep-alive penalty is applied uniformly to all BNAF-based variants that use the shared-trunk critic.

\section{Hyperparameters}
\label{app:MoCA_hparams}

\begin{table}[H]
\centering
\small
\begin{tabular}{lcc}
\toprule
\textbf{Hyperparameter} & \textbf{Cartpole} & \textbf{Pendulum} \\
\midrule
Critic LR (init) & $1\times 10^{-4}$ & $1\times 10^{-4}$ \\
Critic LR (end) & $1\times 10^{-4}$ & $1\times 10^{-4}$ \\
$\tilde{\mu}$ LR (init) & $1\times 10^{-5}$ & $5\times 10^{-6}$ \\
$\tilde{\mu}$ LR (end) & $1\times 10^{-5}$ & $5\times 10^{-6}$ \\
Batch size & 128 & 128 \\
\midrule
$\sigma_\text{explore}$ (init) & 1.0 & 0.3 \\
$\sigma_\text{explore}$ (end) & 0.1 & 0.3 \\
Random-action probability $p_\text{rand}$ & 0.2 & 0.2 \\
\midrule
Target update period $N_\text{target}$ & 100 & 100 \\
\midrule
Curvature keep-alive $\lambda$ & $1\times 10^{-4}$ & $1\times 10^{-4}$ \\
Curvature keep-alive $p_{\min}$ & 0.5 & 2.0 \\
\bottomrule
\end{tabular}
\caption{MoCA hyperparameters used for Cartpole and Pendulum. Learning rates are specified separately for the critic and the quadratic vertex head $\tilde{\mu}(s)$. For Cartpole, exploration noise $\sigma_\text{explore}$ is linearly annealed from its initial to final value over the first 50\% of training steps; for Pendulum, $\sigma_\text{explore}$ is held constant throughout. Curvature keep-alive uses the hinge-squared penalty described in Appendix~\ref{app:curvature_reg}.}
\end{table}
\appendix

\end{document}